\documentclass{article}
\usepackage{graphicx}

\usepackage[preprint]{neurips_2025}

\usepackage[utf8]{inputenc} 
\usepackage[T1]{fontenc}    
\usepackage{hyperref}       
\usepackage{url}            
\usepackage{booktabs}       
\usepackage{amsfonts}       
\usepackage{nicefrac}       
\usepackage{microtype}      
\usepackage{xcolor}         

\usepackage{amsmath, amssymb, amsthm}
\usepackage{geometry}
\usepackage{mathrsfs}
\usepackage{appendix}
\usepackage{graphicx}  
\usepackage{subcaption}
\usepackage{natbib}
\setcitestyle{authoryear,open={[},close={]}}
\hypersetup{pdfborder={0 0 0}, colorlinks=true, citecolor=blue, linkcolor=blue, urlcolor=blue}

\title{Dynamic Perturbed Adaptive Method for Infinite Task-Conflicting Time Series}
\author{
   Jiang YOU \thanks{These authors contributed equally to this work. } \\
   Department of Computer Science\\
   Université Gustave Eiffel \& ESIEE-Paris\\
   \texttt{jiang.you@esiee.fr} \\
   \And
   Xiaozhen WANG $^\ast$ \\
   CEREMADE\\
   Université Paris-Dauphine PSL\\
   \texttt{xiaozhen.wang@dauphine.psl.eu} \\
   \And
   Arben CELA \\
   Department of Computer Science\\
   Université Gustave Eiffel \& ESIEE-Paris\\
  \texttt{arben.cela@esiee.fr} \\
}
\date{April 2025}

\newtheorem{definition}{Definition}
\newtheorem{theorem}{Theorem}

\newtheorem{assumption}{Assumption}

\newtheorem{remark}{Remark}

\begin{document}

\maketitle

\begin{abstract}
    We formulate time series tasks as input-output mappings under varying objectives, where the same input may yield different outputs. This challenges a model’s generalization and adaptability. To study this, we construct a synthetic dataset with numerous conflicting subtasks to evaluate adaptation under frequent task shifts. Existing static models consistently fail in such settings. We propose a dynamic perturbed adaptive method based on a trunk–branch architecture, where the trunk evolves slowly to capture long-term structure, and branch modules are re-initialized and updated for each task. This enables continual test-time adaptation and cross-task transfer without relying on explicit task labels. Theoretically, we show that this architecture has strictly higher functional expressivity than static models and LoRA. We also establish exponential convergence of branch adaptation under the Polyak–Łojasiewicz condition. Experiments demonstrate that our method significantly outperforms competitive baselines in complex and conflicting task environments, exhibiting fast adaptation and progressive learning capabilities.
\end{abstract}


\section{Introduction}


Time series models have achieved strong results in forecasting \citep{wen2022transformers,zhou2021informer}. However, most existing methods rely on a static task formulation, where the input–output relationship is assumed to be consistent across training and deployment. This assumption breaks down in many real-world settings, such as user behavior modeling~\citep{requeima2019fast} or sensor-driven control~\citep{wang2020tent}, where objectives shift over time and the same input can correspond to multiple plausible outputs depending on context or intent\citep{context_matters_2025}.

In this work, we introduce a generalized formulation of time series learning as a sequence of input-output mappings under variable and often conflicting objectives. In this setting, the same input sequence may yield different targets depending on context or task, such as intent, semantics, or environmental conditions. This challenges a model’s ability to generalize beyond average-case behavior and exposes the limitations of static models in dynamic environments~\citep{pham2022learning}. These challenges underscore the need for continual learning frameworks that can adapt on the fly to evolving objectives~\citep{gama2014survey, hosseinzadeh2025efficient} and effectively manage conflicting goals across tasks~\citep{ruder2017overview}.


To evaluate model behavior under shifting objectives, we construct a large-scale synthetic benchmark composed of thousands of mutually conflicting subtasks. All tasks share a compact base of input sequences but differ in output semantics, inducing frequent and structured objective shifts. This design reveals a key limitation of static models: despite their capacity, they collapse toward averaged solutions and fail to adapt under continual contradiction and distributional change.

To address this challenge, we propose a Dynamic Perturbed Adaptive (DPA) framework built upon a trunk–branch architecture. The trunk serves as a slow-evolving representation learner that accumulates long-term structural information across tasks. In contrast, branch modules are re-initialized and fine-tuned on each new task, enabling rapid local adaptation without catastrophic interference. Crucially, this method operates without access to task identifiers, relying instead on a lightweight perturbation-and-adaptation cycle at test time.

We provide theoretical guarantees for our approach: under the Polyak–Łojasiewicz (PL) condition, we establish exponential convergence of the adaptation dynamics. Moreover, we prove that the proposed trunk–branch architecture exhibits strictly higher functional expressivity than both static models and low-rank adaptation (LoRA) approaches, enabling it to represent and interpolate across a broader class of task mappings. Finally, we show that our method achieves sublinear dynamic regret, ensuring robust adaptation in non-stationary and continually shifting task distributions.

Extensive experiments across synthetic benchmarks confirm that our method outperforms state-of-the-art baselines by a wide margin. Notably, it achieves fast test-time adaptation, resilience to task interference, and progressive learning behavior, paving the way toward more robust and flexible time series reasoning models.





We summarize our contributions as follows:
\begin{itemize}
    \item We formulate time series learning as dynamic input-output mappings under shifting and conflicting objectives, and introduce a dynamic perturbed adaptive trunk--branch architecture that supports continual test-time adaptation without task labels.
    \item We provide theoretical guarantees for our approach, including exponential convergence under the Polyak–Łojasiewicz condition and sublinear dynamic regret, and show that our model enjoys higher functional expressivity than static networks and LoRA-style adaptation.
    \item We construct a large-scale synthetic benchmark comprising contradictory task mappings to evaluate model adaptability in highly non-stationary environments.
    \item We empirically demonstrate state-of-the-art performance on complex time series tasks, achieving up to 71.77\% relative error reduction over strong baselines in Table \ref{tab:synthetic_comparison}, including static models and low-rank adaptation methods like LoRA. Our model achieves fast test-time adaptation, robustness to task interference, and continual improvement across dynamic benchmarks.
\end{itemize}

To guide the reader through our approach, we begin in Section 2 with a review of related work on time series adaptation, including parameter-efficient tuning, dynamic modeling, and continual learning. Section 3 introduces a dynamic trunk--branch framework for continual adaptation in shifting time series tasks, with theoretical guarantees on expressivity and convergence. In Section 4, we describe our synthetic benchmark and provide comprehensive experimental results comparing our method to strong baselines. Section 5 discusses the broader implications of our findings and outlines potential limitations. Section 6 outlines the main limitations of our approach, and Section 7 concludes with perspectives on real-world deployment and theoretical extension.

\section{Related Work}

Recent advances in deep learning have been driven by the emergence of large pre-trained models (PTMs), which offer strong generalization across diverse tasks but incur substantial computational costs during downstream adaptation. As model sizes scale to billions of parameters~\citep{brown2020language, chowdhery2023palm}, full fine-tuning becomes increasingly impractical. This has motivated the development of parameter-efficient fine-tuning (PEFT) techniques, which adapt large models by updating only a small subset of parameters or inserting lightweight trainable modules~\citep{houlsby2019parameter, li2021prefix, zaken2021bitfit, hu2021lora}. PEFT reduces memory and compute overhead while supporting faster adaptation, lower data requirements, and improved continual learning.

Time series tasks often involve conflicting objectives—such as simultaneously forecasting rising and falling trends or modeling components with differing temporal dynamics. These challenges call for learning frameworks that can jointly capture shared and task-specific patterns. Multi-task learning (MTL) has been widely adopted to leverage task relatedness~\citep{ruder2017overview}, while continual learning (CL) aims to maintain past knowledge in sequential settings without catastrophic forgetting~\citep{parisi2019continual}.

Beyond task conflict, time series data is inherently non-stationary and susceptible to distributional shifts and concept drift. Test-time training (TTT) enables on-the-fly adaptation to these shifts using unlabeled test data~\citep{sun2020ttt}, while other adaptive strategies—such as online ensembling and drift-aware updates~\citep{gama2014survey}—provide complementary robustness against evolving data distributions.

To accommodate these challenges, recent work has explored PEFT in the context of large-scale foundation models for time series. Techniques such as LoRA~\citep{hu2021lora} introduce low-rank adapters into frozen networks to enable lightweight adaptation, as shown effective in long-horizon forecasting and anomaly detection~\citep{nie2024channel}. TRACE~\citep{li2025trace} extends this idea by dynamically selecting relevant modules and reconfiguring prediction heads to handle temporal non-stationarity. Recent efforts like TimeRAF~\citep{zhang2024timeraf} and TimeFilter \citep{hu2025timefilter} further demonstrate the utility of modular and retrieval-augmented designs for adapting foundation models in dynamic time series environments.

U-Net \citep{ronneberger2015unet}, originally designed for image segmentation, has been effectively adapted for time series tasks due to its encoder–decoder structure with skip connections, which captures both short- and long-range temporal dependencies. Time series variants typically adopt 1D convolutions and are used for forecasting and anomaly detection \citep{you2024kernel, dong2024mai}. 



Recent work has extended U-Net beyond its original use in image segmentation to time series tasks by incorporating mechanisms such as attention, dynamic kernels, and customized receptive fields to better capture non-stationary and multi-scale temporal patterns \citep{cui2024attention,alvarado2025sdku}. In particular, Kernel U-Net \citep{you2024kernel} decouples the partitioning of time series inputs from kernel execution, enabling flexible and task-specific kernel designs. This modularity makes it well-suited for analyzing dynamic task behaviors and highlights U-Net's adaptability as a general-purpose backbone for time series modeling.

While recent PEFT frameworks such as LoRA and TRACE have improved adaptation efficiency in time series foundation models, they remain limited by fixed shared backbones and lack fine-grained task-level flexibility. These constraints hinder performance in dynamic settings with frequent distribution shifts and conflicting objectives. To address this, we propose a dynamic perturbed adaptive trunk--branch architecture that separates stable long-term representations from rapidly re-initialized, task-adaptive modules, enabling continual adaptation without requiring explicit task supervision.

\section{Method: A Dynamic Perturbed Adaptive Trunk--Branch Architecture}

\begin{figure}[htbp]
    \centering
    \includegraphics[width=0.95\linewidth]{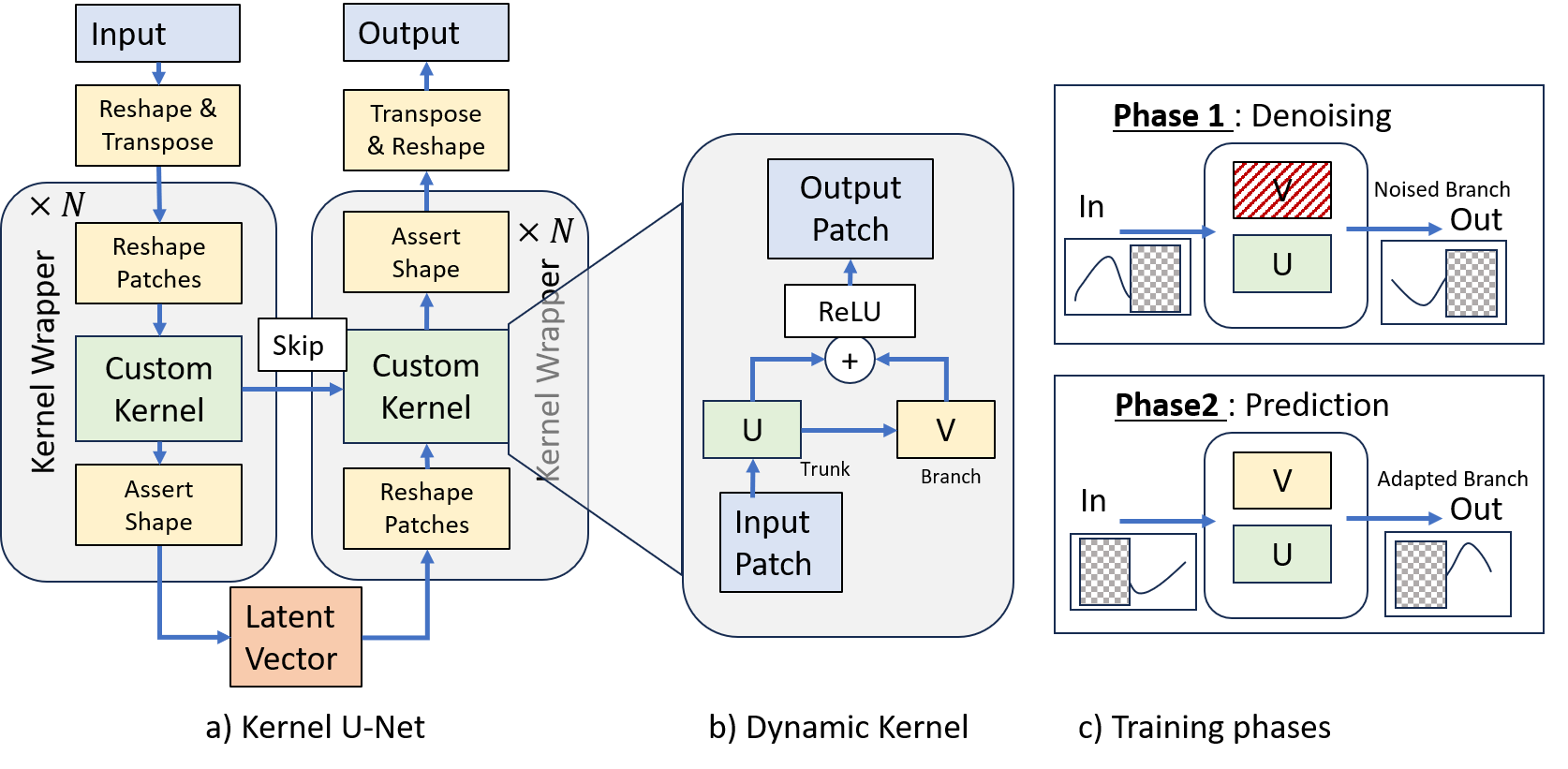}
    \caption{a) Illustration of Kernel U-Net. b) We note $U$  the Trunk and $V$ is the Branch for constructing Dynamic Kernel. Trunk evolves slowly through all tasks, and the Branch is reinitialized at each new task. c) In detail when training a dynamic kernel, the phase 1 allows the Branch to adapt the task context, and the pahse 2 allows the model predicting the following values.}
    \label{fig:illustration_dynamic_kernel_unet}
\end{figure}

Real-world time series tasks often involve rapidly shifting objectives~\citep{finn2017model}, where the same input may yield different targets depending on context~\citep{requeima2019fast}. This challenges a model’s ability to generalize across tasks and adapt swiftly to new patterns. Static networks, with fixed parameters, are fundamentally limited in such settings. Even low-rank adaptation methods like LoRA~\citep{hu2021lora}, though offering some flexibility, rely on a shared base and do not support continual task-level reinitialization.

To address this, we propose a \textbf{dynamic perturbed adaptive trunk--branch architecture} that combines long-term stability with short-term adaptability. The \emph{trunk} comprises layers updated slowly via small cumulative perturbations, forming a stable backbone for cross-task representations. The \emph{branch modules} are lightweight, re-initialized per task, and optimized rapidly without requiring task identifiers. This design enables adaptation to a stream of conflicting subtasks and frequent shifts. Unlike prior methods based on fixed task heads or continuous fine-tuning~\citep{nichol2018first,wang2020tent}, our approach supports continual test-time adaptation through a structured separation of dynamics.

Formally, we define a depth-\(N\) network with \(N = L + M\), arranged as a trunk–branch tree. Let \(x_t \in \mathbb{R}^{d_0}\) be the input and \(y_t \in \mathbb{R}^{m}\) the output at time \(t\). The network has layer dimensions \((d_0, \dots, d_N)\), with each layer mapping \(\mathbb{R}^{d_\ell} \to \mathbb{R}^{d_{\ell+1}}\). Trunk layers \(\ell = 0,\dots,L-1\) apply
\[
U^{(\ell)}_{\Phi^{(\ell)}_t} : \mathbb{R}^{d_\ell} \to \mathbb{R}^{d_{\ell+1}}, \quad \Phi^{(\ell)}_t = \Phi^{(\ell)} + \Theta^{(\ell)}_t,
\]
with fixed base \(\Phi^{(\ell)}\) and time-varying perturbation \(\Theta^{(\ell)}_t \in \mathbb{R}^{p_\ell}\). Branch layers \(\ell = L,\dots,N-1\) apply
\[
V^{(\ell)}_{\Psi^{(\ell)}_t} : \mathbb{R}^{d_\ell} \to \mathbb{R}^{d_{\ell+1}}, \quad \Psi^{(\ell)}_t \sim \mathcal{I}_\ell,
\]
with \(\Psi^{(\ell)}_t \in \mathbb{R}^{q_\ell}\) freshly initialized from distribution \(\mathcal{I}_\ell \in \mathbb{R}^{q_\ell}\) and trained locally. The full model output is
\begin{equation}
\label{eq:full_model}
    y_t = \big( V^{(N-1)}_{\Psi^{(N-1)}_t} \circ \cdots \circ V^{(L)}_{\Psi^{(L)}_t} \big) \circ \big( U^{(L-1)}_{\Phi^{(L-1)}_t} \circ \cdots \circ U^{(0)}_{\Phi^{(0)}_t}(x_t) \big).    
\end{equation}

At each step \(t\), the model is adapted to the current task by jointly optimizing branch parameters \(\Psi_t = \{\Psi^{(\ell)}_t\}_{\ell=L}^{N-1}\) and trunk perturbations \(\Theta_t = \{\Theta^{(\ell)}_t\}_{\ell=0}^{L-1}\). The loss is defined as
\[
\mathcal{L}^{(t)} = \ell(y_t,\,y^{\mathrm{true}}_t) + \sum_{\ell=0}^{L-1} \gamma_\ell \|\Theta^{(\ell)}_t\|^2,
\]
where \(\ell(\cdot,\cdot)\) is task-specific (e.g., MSE or cross-entropy), and the regularization penalizes excessive trunk perturbations. Branches are updated using fixed learning rate \(\beta\) over a short inner loop:
\begin{equation}
\label{eq:Psi_process}
    \Psi^{(\ell)}_{t+1} = \Psi^{(\ell)}_t - \beta \nabla_{\Psi^{(\ell)}_t}\, \mathcal{L}^{(t)}, \quad \ell = L, \dots, N-1,
\end{equation}
while trunk updates use smaller, layerwise learning rates:
\begin{equation}
\label{eq:Theta_process}
    \Theta^{(\ell)}_{t+1} = \Theta^{(\ell)}_t - \alpha_\ell(t) \nabla_{\Theta^{(\ell)}_t}\, \mathcal{L}^{(t)}, \quad \ell = 0, \dots, L-1,
\end{equation}
with \(0 < \alpha_0(t) \le \cdots \le \alpha_{L-1}(t) \ll \beta\). Branches are reset at each task, while trunk parameters accumulate knowledge over time.

\subsection{Expressivity of the Trunk–Branch Architecture}
Frequent task shifts require models to approximate diverse functions across episodes. Static networks, with fixed parameters, must compromise globally and incur high worst-case error. Our dynamic trunk–branch design addresses this by separating roles: the trunk accumulates shared structure over time, while branches adapt locally through per-task reinitialization. 

We formalize this by defining the static hypothesis space \(\mathcal{H}_{\mathrm{static}}\) and the dynamic hypothesis space \(\mathcal{H}_{\mathrm{dyn}}\) below. Let \(\mathcal{X} \subset \mathbb{R}^{d_0}\) be the input space and \(\mathcal{Y} \subset \mathbb{R}^m\) the output space. Fix \(N = L + M\) layers. For each trunk layer \(\ell = 0, \dots, L-1\), let \(\Theta^{(\ell)}_t \in \mathbb{R}^{p_\ell}\) as defined in~\eqref{eq:Theta_process}; for each branch layer \(\ell = L, \dots, N-1\), let \(\Psi^{(\ell)}_t \in \mathbb{R}^{q_\ell}\) as in~\eqref{eq:Psi_process}. The total parameter budget is
\[
P := \sum_{\ell=0}^{L-1} p_\ell + \sum_{\ell=L}^{N-1} q_\ell.
\]

\paragraph{Static Model.} Define the static hypothesis space as
\begin{equation}
\label{eq:H_static}
    \mathcal{H}_{\mathrm{static}} = \Bigl\{
    f(x) = \big( V^{(N-1)}_{\Psi^{(N-1)}} \circ \cdots \circ V^{(L)}_{\Psi^{(L)}}\big) \circ \big(U^{(L-1)}_{\Phi^{(L-1)}} \circ \cdots \circ U^{(0)}_{\Phi^{(0)}}\big)(x)
    \;\Big|\; \text{all parameters fixed}
    \Bigr\}.
\end{equation}

\paragraph{Dynamic Model.} 
Let \(\mathcal{F}_t\) denote the set of functions expressible by the model at time \(t\), with trunk parameters \(\Phi^{(\ell)}_t = \Phi^{(\ell)} + \Theta^{(\ell)}_t\) and branch parameters \(\Psi_t = \{\Psi_t^{(\ell)}\}_{\ell=L}^{N-1}\). Then
\[
\mathcal{F}_t = \left\{
f_t(x) = V^{(N-1)}_{\Psi^{(N-1)}_t} \circ \cdots \circ V^{(L)}_{\Psi^{(L)}_t} \circ U^{(L-1)}_{\Phi^{(L-1)}_t} \circ \cdots \circ U^{(0)}_{\Phi^{(0)}_t}(x)
\;\middle|\;
\Phi^{(\ell)}_t \in \mathbb{R}^{p_\ell},
\Psi^{(\ell)}_t \in \mathbb{R}^{q_\ell}
\right\}.
\]
Each \(f_t \in \mathcal{F}_t\) represents a function composed from the current branch and the perturbed trunk. The dynamic hypothesis space is defined as the union over time:
\begin{equation}
\label{eq:H_dyn}
    \mathcal{H}_{\mathrm{dyn}} = \bigcup_{t \in [0,T]} \mathcal{F}_t \;\subset\; L^2(\mathcal{X}, \mathcal{Y}).
\end{equation}

To quantify expressivity, we use the Kolmogorov \(n\)-width~\citep{pinkus2012n}, which measures the best possible \(L^2\) approximation error achievable by any \(n\)-dimensional subspace.

\begin{definition}[Kolmogorov \(n\)-width]
Let \(\mathcal{F} \subset L^2(\mathcal{X}, \mathcal{Y})\) be bounded. The \(n\)-width of \(\mathcal{F}\) is
\[
d_n(\mathcal{F}) := \inf_{\dim V \le n} \sup_{f \in \mathcal{F}} \inf_{g \in V} \|f - g\|_{L^2}.
\]
This quantity measures the smallest worst‐case $L^2$‐approximation error of $\mathcal F$ by any $n$‐dimensional subspace.
\end{definition}
\begin{theorem}[Width Reduction via Time Partitioning]
\label{thm:expressivity}
Let \(\mathcal{H}_{\mathrm{static}}\) and \(\mathcal{H}_{\mathrm{dyn}}\) be as defined above, with total parameter budget \(P\). Suppose we partition \([0,T]\) into \(K\) episodes, and for each episode \(i = 1,\dots,K\), define
\[
\mathcal{F}_i = \bigcup_{t \in [t_{i-1}, t_i)} \mathcal{F}_t.
\]
Let \(n_i\) be the parameter allocation per episode, with \(\sum_{i=1}^K n_i = P\). Then
\[
d_P(\mathcal{H}_{\mathrm{dyn}}) \le d_P(\mathcal{H}_{\mathrm{static}}),
\]
with strict inequality whenever \(\max_i d_{n_i}(\mathcal{F}_i) < d_P(\mathcal{H}_{\mathrm{static}})\).
\end{theorem}
\noindent See Appendix~\ref{proof:expressivity_proof} for the proof. The Kolmogorov \(n\)-width \(d_n(\mathcal{F})\) quantifies the minimal worst-case approximation error achievable by any \(n\)-dimensional subspace. By partitioning time and reinitializing branches across episodes, our model constructs a union of specialized subspaces, enabling significantly tighter approximation for the same parameter budget. This improvement stems from the explicit separation of dynamics: a slowly evolving trunk captures persistent structure, while freshly reset branches adapt sharply to local variations. In contrast, static networks, LoRA-style low-rank adapters, and online fine-tuning lack either structural decoupling or episodic flexibility—limiting their functional coverage. Our trunk–branch design overcomes both limitations, achieving provably greater expressivity under dynamic workloads.

\begin{remark}[Optimal Allocation]
Suppose each segment-wise function class \(\mathcal{F}_i\) lies in a Sobolev ball \(W_2^s(\Omega)\), the space of functions on \(\Omega \subset \mathbb{R}^{d_0}\) whose weak derivatives up to order \(s\) are square-integrable. Then its Kolmogorov width satisfies
\[
d_{n_i}(\mathcal{F}_i)\; \le\; C\,n_i^{-s/d_0}.
\]
Here, the smoothness parameter \(s\) characterizes the regularity of the target functions: higher \(s\) corresponds to smoother functions with bounded high-order derivatives, which are easier to approximate; lower \(s\) indicates less regular functions that require greater capacity to represent. Under a uniform allocation \(n_i = P/T\), we obtain:
\[
d_P(\mathcal{H}_{\mathrm{dyn}})
\;\le\;
\max_i\, d_{n_i}(\mathcal{F}_i)
\;\le\;
C\bigl(P/T\bigr)^{-s/d_0}
=
T^{s/d_0} \cdot C\,P^{-s/d_0}
=
T^{s/d_0} \cdot d_P(\mathcal{H}_{\mathrm{static}}).
\]
Hence, the hierarchical dynamic model reduces the worst-case approximation error by a factor of \(T^{-s/d_0}\),  
or equivalently, amplifies expressivity by a factor of \(T^{s/d_0}\), compared to any static model of the same size.
\end{remark}


\subsection{Convergence of trunk Updates}
\label{sec:convergence}
The dynamic trunk–branch architecture enables continual test-time adaptation by incrementally updating the trunk while resetting branches for each task. To ensure stable long-term learning under non-stationary objectives, we analyze the convergence of the trunk parameters \(\Theta_t = \{\Theta_t^{(\ell)}\}_{\ell=0}^{L-1}\). Specifically, we show that under the Polyak–Łojasiewicz (PL) condition, which captures a broad class of overparameterized and nonconvex problems \citep{karimi2016linear,allen2019convergence}, the sequence \(\Theta_t\) converges to the global minimizer of the time-averaged loss.

We define the time-averaged trunk objective as
\begin{equation}
\label{eq:avg_loss}
F(\Theta) := \lim_{T \to \infty} \frac{1}{T} \sum_{t=0}^{T-1} \mathcal{L}^{(t)}(\Theta).
\end{equation}

We now state the assumptions under which convergence holds.

\begin{assumption}[Smoothness and PL Inequality]
\label{assump:PL}
Let \(F(\Theta)\) be defined in~\eqref{eq:avg_loss}. We assume:
\begin{itemize}
    \item \(\mathcal{L}^{(t)}\) is \(L\)-smooth for all \(t\), i.e., $\|\nabla \mathcal{L}^{(t)}(\Theta) - \nabla \mathcal{L}^{(t)}(\Theta')\| \le L \|\Theta - \Theta'\|, \quad \forall\, \Theta, \Theta'$.
    \item PL Inequality: There exists \(\mu > 0\) such that for all \(\Theta\), $\frac{1}{2} \|\nabla F(\Theta)\|^2 \ge \mu \big(F(\Theta) - F^*\big)$, where $F^* = \min_\Theta F(\Theta)$.
\end{itemize}
\end{assumption}

\begin{assumption}[Bounded gradient drift]
\label{assump:drift}
There exists a sequence \(\{\delta_t\} \in \mathbb{R}_+\) with \(\sum_t \delta_t < \infty\), such that for all \(t\),
\[
\big\| \nabla \mathcal{L}^{(t)}(\Theta) - \nabla F(\Theta) \big\| \le \delta_t.
\]
\end{assumption}
We now establish that the hierarchical trunk updates converge linearly to the global minimum under these assumptions.
\begin{theorem}[Linear Convergence under PL Condition]
\label{thm:PL_convergence}
Let the trunk parameters \(\Theta_t = \{\Theta_t^{(\ell)}\}_{\ell=0}^{L-1}\) be updated according to~\eqref{eq:Theta_process}, with step sizes satisfying
\[
0 < \alpha_0(t) \le \cdots \le \alpha_{L-1}(t) < \tfrac{1}{L}, \quad
\underline{\alpha} := \inf_t \alpha_0(t) > 0.
\]

Suppose Assumptions~\ref{assump:PL} and~\ref{assump:drift} hold, and that the drift-weighted conditions
\[
\sum_{t=0}^\infty \alpha_0(t)\,\delta_t < \infty,
\quad
\sum_{t=0}^\infty \alpha_0(t)^2\,\delta_t^2 < \infty
\]
are satisfied. These conditions are standard in stochastic approximation and online optimization to ensure convergence under non-stationary noise~\citep{bottou2018optimization,mairal2013stochastic}.

Then the time-averaged objective satisfies the contraction
\begin{equation*}
    F(\Theta_{t+1}) - F^* \le \big( 1 - \mu \alpha_0(t)\big) \big(F(\Theta_t\big) - F^*) + \alpha_0(t)\delta_t + \tfrac{L}{2}\alpha_0(t)^2 \delta_t^2.
\end{equation*}

In particular, if \(\underline{\alpha} > 0\), then
\[
F(\Theta_t) - F^* \le (1 - \mu \underline{\alpha})^t (F(\Theta_0) - F^*) + o(1),
\]
achieving linear convergence to the global minimizer \(F^*\).
\end{theorem}

\noindent See Appendix~\ref{proof:PL_proof} for a complete proof. This result ensures that despite frequent task resets in the branch, the trunk parameters evolve smoothly toward a shared structure, enabling stable long-term accumulation across tasks.

\subsection{Dynamic Regret}
In non-stationary settings with continually shifting objectives, static convergence guarantees are insufficient. A standard measure of adaptability in such online regimes is the \emph{dynamic regret}, defined as
\begin{equation}
\label{eq:dynamic_regret}
R_T := \sum_{t=0}^{T-1} \left[ \mathcal{L}^{(t)}(\Theta_t) - \mathcal{L}^{(t)}(\Theta_t^*) \right],
\end{equation}
where \(\Theta_t^* := \arg\min_\Theta \mathcal{L}^{(t)}(\Theta)\) denotes the instantaneous optimal trunk parameter at time \(t\). We aim to show that our trunk–branch updates maintain \emph{sublinear} dynamic regret, ensuring that the average regret per time step vanishes asymptotically.

\begin{assumption}[Smoothness and Bounded Gradients]
\label{assump:DR}
Each \(\mathcal{L}^{(t)}(\cdot)\) is \(L\)-smooth, and the gradient norms are uniformly bounded: \(\|\nabla \mathcal{L}^{(t)}(\Theta)\| \le G\) for all \(\Theta\) and \(t\).
\end{assumption}

\begin{theorem}[Sublinear Dynamic Regret]
\label{thm:dynamic_regret}
Under Assumption~\ref{assump:DR}, the trunk updates defined in~\eqref{eq:Theta_process} satisfy
\[
R_T = \mathcal{O}\left( \sum_{t=0}^{T-1} \alpha_{L-1}(t) + \frac{CV_T}{\min_t \alpha_0(t)} \right),
\]
where \(CV_T := \sum_{t=1}^{T-1} \| \Theta_t^* - \Theta_{t-1}^* \|\) denotes the cumulative variation of the optimal trunk path.

In particular, if the step sizes are fixed as \(\alpha_0(t) = \alpha_{L-1}(t) = \mathcal{O}(1/\sqrt{T})\), then
\[
R_T = \mathcal{O}\left( \sqrt{T} + CV_T \cdot \sqrt{T} \right),
\]
which implies vanishing average regret as \(T \to \infty\), provided \(CV_T = o(\sqrt{T})\).
\end{theorem}

See Appendix~\ref{app:dynamic_regret} for proof details. This result ensures that the slowly evolving trunk can track non-stationary targets while preserving long-term structure, thus enabling continual generalization even in highly dynamic environments.

\section{Dataset}

To evaluate the adaptability of models in multi-task scenarios under controlled conditions, we construct a synthetic dataset composed of sine wave sequences with systematic variations. This dataset allows precise control over task structure and complexity, facilitating the analysis of model behavior during task switching and adaptation.

\subsection{Data Generation Protocol}

The core elements of the dataset are sine sequences with a base period of $T_0 =3$. Each sequence undergoes a random horizontal shift and amplitude scaling. From these continuous sequences, we extract a set of sub-sequences with fixed input and output lengths. These sub-sequences are selected using a sliding window approach with controlled boundaries to ensure consistency in sample structure.

During sampling, the algorithm proceeds as follows:
\begin{itemize}
    \item Randomly select an index $i$ corresponding to a specific sine base sequence.
    \item Extract 10 consecutive sub-sequences starting from index $i$, each with a fixed length. The first 5 are used as training examples, while the remaining 5 form the test set.
    \item Independently sample another index $j$, and randomly pair it with $i$ such that each sub-sequence in the input (index $i$) is matched with a corresponding output from index $j$.
\end{itemize}

This setup forces the model to rapidly infer the task mapping from a small number of input-output training pairs, and then generalize to predict outputs for the test inputs. The goal is to simulate dynamic task inference, where task identity is implicitly encoded in the input-output examples rather than provided explicitly.

\subsection{Dataset Variants}

To evaluate the model under increasingly challenging settings, we design three versions of the dataset:
\begin{itemize}
    \item \textbf{Dataset S1 (Simple Sine):} Composed of 5 pure sine wave segments with random amplitude and phase shifts but identical period $T_0=3$.
    \item \textbf{Dataset S2 (Mixed Amplitude and Phase):} Each sequence is the sum of two sine waves with the same period but different amplitudes and phase shifts, introducing interference and amplitude modulation.
    \item \textbf{Dataset S3 (Mixed Periods):} Each sequence is generated by summing two sine waves with different periods, amplitudes, and phase shifts, resulting in more complex temporal patterns and increased task difficulty.
\end{itemize}
This design enables the systematic study of model adaptation dynamics as the complexity of the underlying temporal structure increases.

\begin{figure*}[htbp]
\centering
\begin{subfigure}[t]{0.32\textwidth}
    \centering
    \includegraphics[width=\linewidth]{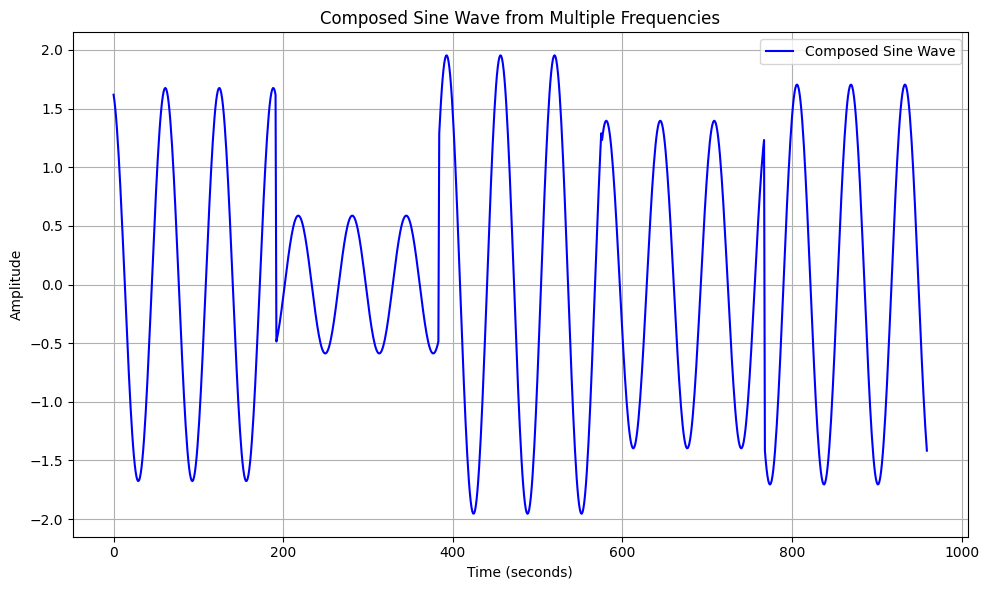}
    \caption{Dataset S1}
    \label{fig:mse_s1}
\end{subfigure}
\hfill
\begin{subfigure}[t]{0.32\textwidth}
    \centering
    \includegraphics[width=\linewidth]{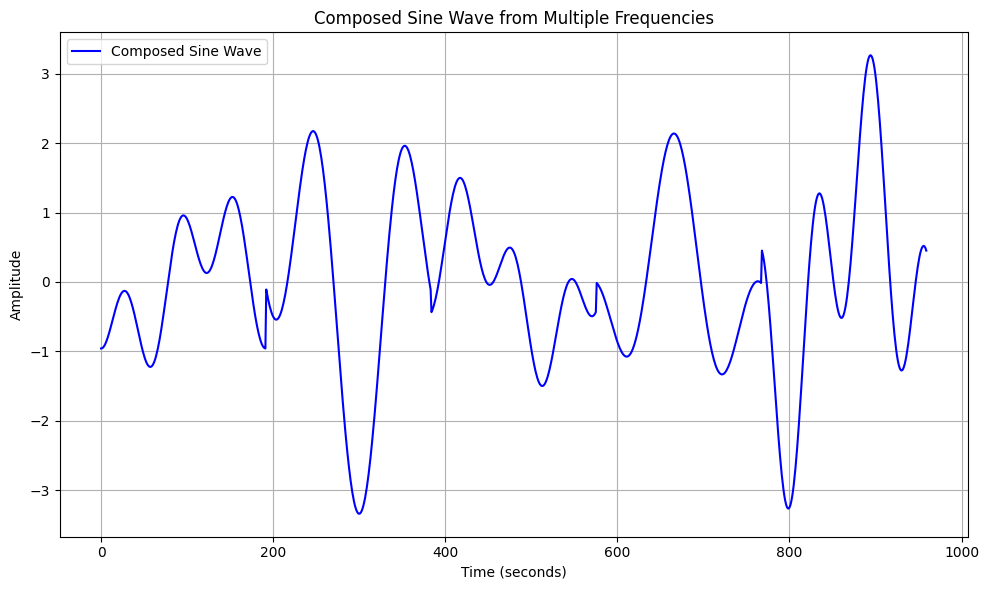}
    \caption{Dataset S2}
    \label{fig:mse_s2}
\end{subfigure}
\hfill
\begin{subfigure}[t]{0.32\textwidth}
    \centering
    \includegraphics[width=\linewidth]{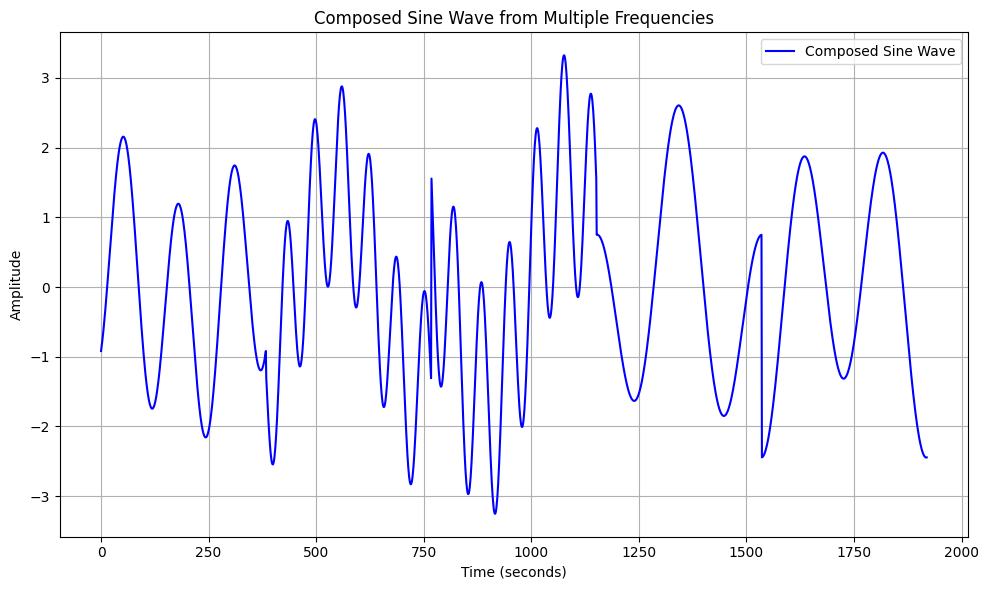}
    \caption{Dataset S3}
    \label{fig:mse_s3}
\end{subfigure}
\caption{
Illustation of synthetic dataset : (a) Dataset S1; (b) Dataset S2; (c) Dataset S3.
}
\label{fig:ablation_three_plots}
\end{figure*}

\section{Experiment Design}
To evaluate how neural networks behave when exposed to noisy, contradictory mappings, we design an experiment that combines controlled synthetic data with test-time adaptation strategies. Our goal is not only to observe convergence trends under traditional learning settings, but also to assess how well a model can dynamically adapt when no stable input-output relationship exists.

Each experiment consists of multiple iterations. In every iteration, we initialize a fresh model and train it from scratch using a synthetic dataset composed of randomly assembled input-output subsequences. The model is optimized using the Adam optimizer with a fixed learning rate, and the Mean Squared Error (MSE) loss is used to measure performance across training, validation, and testing.

Training proceeds in epochs, with each epoch consisting of a limited number of batches for computational efficiency. For each batch, the training procedure is divided into two stages. First, a test-time training phase is applied: the model is updated using the first five subsequences in the batch, which serve as input-output pairs to compute gradients for a latent adaptation mechanism. This stage simulates how the model might adapt during testing in real-world scenarios, where conditions can change, and the model must adjust without additional supervision. A custom gradient modification function is applied after backpropagation to alter internal representations. 

After the branch adaptation stage, the model is evaluated on the following five subsequences of the same batch, which act as a test sample. The loss from this stage is used to update the model weights in the main prediction phase. The first stage is  executed for 10 steps in an inner loop iteration to enable sufficient adaptation and stabilization of parameters in the branch. In the second stage, the model executes only once for training the parameters in the trunk or for evaluation without updating parameters. 

For this experiment, we set the number of epochs to 50 to ensure sufficient training iterations for model convergence. The adaptation phase is configured with a total of 10 steps in general or 30 steps in the fine-tuning stage, allowing the model to gradually adjust its parameters. The branch learning rate is set to $\beta = 1e-3$, while the trunk learning rate is set to $\alpha_{\ell}(t) = 3e-5$ to balance the learning dynamics of the different components. The model architecture uses Kernel U-Net stacked in two layers, incorporating LayerNorm within the network for stable training. The latent dimension is set to 128, providing a rich feature space for representation learning. The training is conducted on a GPU (NVIDIA L4) on Google Colab, ensuring efficient computation throughout the process.

\begin{table}[htbp]
\centering
\resizebox{\textwidth}{!}{%
\begin{tabular}{|l|c|cc|c|ccc|ccc|}
\hline
 & & \multicolumn{2}{c|}{\textbf{Dynamic}} & \textbf{LoRA}  & \multicolumn{3}{c|}{\textbf{Init-All}} & \multicolumn{3}{c|}{\textbf{Static}} \\
\cline{3-11}
\textbf{Model} & \textbf{Imp ↓} & \textbf{K-U-Net *} & \textbf{K-U-Net} & \textbf{K-U-Net} & \textbf{K-U-Net} & \textbf{PatchTST} & \textbf{Linear} & \textbf{K-U-Net} & \textbf{PatchTST} & \textbf{Linear} \\
\hline
S-1 & 71.77\% & \textbf{0.0858} & 0.2205 & \underline{0.3039} & 0.552 & 1.3406 & 1.191 & 0.8905 & 0.9296 & 1.0401 \\

S-2 & 43.64\% & \textbf{0.57} & 0.8755 & 1.2239 & 1.504 & 1.4553 & 1.0884 & 1.3175 & 1.1493 & \underline{1.0113} \\

S-3 & 46.46\% & \textbf{0.567} & 1.0351 & 1.14 & 1.175 & 1.5166 & 1.3707 & 1.1761 & \underline{1.059} & 1.1527 \\
\hline
\end{tabular}
}

\vspace{0.3em}
\caption{Performance comparison of models across synthetic datasets. * Star marks the version of expeirment with K-U-Net uses 30 epochs of adaption instead of using 10 epochs in the other experiments.}
\label{tab:synthetic_comparison}
\end{table}


The results in Table \ref{tab:synthetic_comparison} highlight the superior performance of the optimized K-U-Net (*K-U-Net), particularly in terms of minimizing mean squared error (MSE) across synthetic datasets (S-1, S-2, and S-3). The optimized version of K-U-Net exhibits the lowest MSE values, achieving 0.0858 for S-1, indicating its ability to adapt effectively to dynamic task switching. This adaptation capability is crucial in time series forecasting tasks, where the model is required to quickly adjust to different patterns across tasks. In comparison, static models like K-U-Net, PatchTST, and Linear show higher MSE values (0.2205, 0.552, and 1.3406, respectively for S-1), suggesting they are less flexible in responding to quick task changes, especially in dynamic environments. LoRA-based configurations show relatively worse performance, especially in S-1 where the error rate is notably higher (0.3039), indicating that the dynamic optimization of K-U-Net provides a substantial advantage.


As dataset complexity increases (S-2 and S-3), the optimized K-U-Net continues to outperform baselines, achieving MSEs of 0.57 and 0.567. In contrast, static models like PatchTST and Linear degrade significantly (e.g., 1.504 and 1.175 on S-2 and S-3). LoRA-based models offer modest improvements over static ones but still underperform (1.2239 and 1.14), highlighting their limited adaptability. These results underscore the robustness of our dynamic architecture in rapidly changing environments. For visualizations and detailed comparisons, see Appendix~\ref{appendix-figure-result}.



\section{Limitations}
While our proposed framework provides a new perspective on evaluating time series models under conflicting task conditions, it remains a synthetic benchmark. The dataset, although diverse, is constructed from controlled base sequences and may not fully capture the complexity and ambiguity found in real-world temporal data. Moreover, our adaptive reasoning method relies on perturbation-based self-adjustment, which, while effective, introduces additional computational overhead and lacks formal theoretical guarantees for convergence or stability in highly noisy settings

\section{Conclusions and future work}
This work reveals the limitations of existing static time series models when facing mutually exclusive or contradictory tasks. By designing a challenging benchmark with structurally incompatible task mappings, we demonstrate that conventional approaches fail to generalize or adapt. Our proposed dynamically perturbed reasoning framework addresses this gap by enabling continuous self-adjustment at test time, achieving superior performance across a wide range of tasks without explicit task supervision. These findings suggest that adaptability, rather than static generalization, is crucial in environments with ambiguous or evolving task definitions.

In future research, we plan to extend our framework to incorporate real-world datasets with implicit task ambiguity, such as multi-intent user behavior logs or medical monitoring data with overlapping diagnoses. We also aim to investigate more theoretically grounded adaptation mechanisms that can provide robustness guarantees and improve efficiency. Additionally, exploring hybrid architectures that combine memory-based retrieval with dynamic perturbation may further enhance long-term task consistency and enable meta-level reasoning across task histories.

\bibliographystyle{plainnat}
\bibliography{references.bib}

\clearpage
\appendix
\section{Proof of Theorem~\ref{thm:expressivity}}
\label{proof:expressivity_proof}

\begin{proof}
By definition, the dynamic hypothesis space is the union over time:
\[
  \mathcal{H}_{\mathrm{dyn}} = \bigcup_{t \in [0,T]} V_t = \bigcup_{i=1}^K \mathcal{F}_i,
\]
where \(\mathcal{F}_i := \bigcup_{t \in [t_{i-1}, t_i)} V_t\) denotes the function class active in segment \(i\). 

By Theorem~1.4.2 in~\citep{pinkus2012n}, the Kolmogorov width of a finite union satisfies
\[
  d_P\left( \bigcup_{i=1}^K \mathcal{F}_i \right) \le \max_{1 \le i \le K} d_{n_i}(\mathcal{F}_i),
\]
for any allocation \((n_1,\dots,n_K)\) with \(\sum_i n_i = P\). 

In particular, taking the degenerate allocation \(n_1 = P\), \(n_{i \ge 2} = 0\), recovers the static model as a special case:
\[
\max_i d_{n_i}(\mathcal{F}_i) \le d_P(\mathcal{H}_{\mathrm{static}}),
\]
which establishes the non-strict inequality.

If, in addition, \(\max_i d_{n_i}(\mathcal{F}_i) < d_P(\mathcal{H}_{\mathrm{static}})\), then
\[
  d_P(\mathcal{H}_{\mathrm{dyn}}) < d_P(\mathcal{H}_{\mathrm{static}}),
\]
which proves the strict gain in expressivity.
\end{proof}

We now turn to proving the convergence guarantees of the hierarchical updates.

\section{Proof of Theorem~\ref{thm:PL_convergence}}
\label{proof:PL_proof}

\begin{proof}
By definition~\eqref{eq:avg_loss}, the time-averaged trunk objective is
\[
F(\Theta) = \lim_{T \to \infty} \frac{1}{T} \sum_{t=0}^{T-1} \mathcal{L}^{(t)}(\Theta),
\]
and Assumption~\ref{assump:PL} guarantees that \(F\) is \(L\)-smooth and satisfies the Polyak--Łojasiewicz (PL) inequality:
\[
\frac{1}{2} \|\nabla F(\Theta)\|^2 \ge \mu\left(F(\Theta) - F^*\right), \quad \forall\, \Theta.
\]

Let \(A(t)\) denote the block-diagonal matrix with blocks \(\alpha_\ell(t)\, \mathrm{Id}_{p_\ell}\), so that the update in~\eqref{eq:Theta_process} becomes:
\[
\Theta_{t+1} = \Theta_t - A(t) \nabla \mathcal{L}^{(t)}(\Theta_t).
\]
By \(L\)-smoothness of \(F\), we have
\[
F(\Theta_{t+1}) \le F(\Theta_t) 
- \langle \nabla F(\Theta_t), A(t) \nabla \mathcal{L}^{(t)}(\Theta_t) \rangle
+ \frac{L}{2} \|A(t) \nabla \mathcal{L}^{(t)}(\Theta_t)\|^2.
\]

Decompose the gradient as
\[
\nabla \mathcal{L}^{(t)}(\Theta_t)
= \nabla F(\Theta_t) + \left(\nabla \mathcal{L}^{(t)}(\Theta_t) - \nabla F(\Theta_t)\right),
\]
and apply Assumption~\ref{assump:drift}, which bounds the drift by
\[
\|\nabla \mathcal{L}^{(t)}(\Theta_t) - \nabla F(\Theta_t)\| \le \delta_t.
\]
It follows that
\[
\|A(t) \nabla \mathcal{L}^{(t)}(\Theta_t)\| \le \alpha_0(t) \|\nabla F(\Theta_t)\| + \alpha_0(t)\delta_t,
\]
and by Cauchy--Schwarz,
\[
\left| \langle \nabla F(\Theta_t), A(t)\left( \nabla \mathcal{L}^{(t)}(\Theta_t) - \nabla F(\Theta_t) \right) \rangle \right|
\le \alpha_0(t) \|\nabla F(\Theta_t)\| \delta_t.
\]

Substituting back, we obtain:
\[
\begin{aligned}
F(\Theta_{t+1}) - F(\Theta_t)
&\le -\alpha_0(t) \|\nabla F(\Theta_t)\|^2
+ \alpha_0(t) \|\nabla F(\Theta_t)\| \delta_t \\
&\quad + \frac{L}{2} \left( \alpha_0(t) \|\nabla F(\Theta_t)\| + \alpha_0(t) \delta_t \right)^2 \\
&= -\alpha_0(t) \|\nabla F(\Theta_t)\|^2
+ \alpha_0(t) \|\nabla F(\Theta_t)\| \delta_t \\
&\quad + \frac{L}{2} \alpha_0(t)^2 \|\nabla F(\Theta_t)\|^2
+ L \alpha_0(t)^2 \|\nabla F(\Theta_t)\| \delta_t
+ \frac{L}{2} \alpha_0(t)^2 \delta_t^2.
\end{aligned}
\]

Since \(\alpha_0(t) < \tfrac{1}{L}\), we bound
\[
-\alpha_0(t) \|\nabla F\|^2 + \frac{L}{2} \alpha_0(t)^2 \|\nabla F\|^2 
\le -\tfrac{1}{2} \alpha_0(t) \|\nabla F\|^2.
\]
Thus,
\[
F(\Theta_{t+1}) - F(\Theta_t)
\le -\tfrac{1}{2} \alpha_0(t) \|\nabla F(\Theta_t)\|^2
+ O\left(\alpha_0(t) \delta_t + \alpha_0(t)^2 \delta_t^2\right).
\]

Applying the PL inequality again:
\[
\|\nabla F(\Theta_t)\|^2 \ge 2\mu (F(\Theta_t) - F^*),
\]
we conclude that
\[
F(\Theta_{t+1}) - F^* \le (1 - \mu \alpha_0(t)) (F(\Theta_t) - F^*) 
+ \alpha_0(t) \delta_t + \tfrac{L}{2} \alpha_0(t)^2 \delta_t^2.
\]

Under the conditions
\[
\sum_{t=0}^\infty \alpha_0(t)\, \delta_t < \infty, \quad
\sum_{t=0}^\infty \alpha_0(t)^2\, \delta_t^2 < \infty,
\]
which are standard for convergence under gradient drift~\citep{bottou2018optimization,mairal2013stochastic}, the additive error vanishes asymptotically. The multiplicative contraction \(\prod_{i=0}^{t-1}(1 - \mu \alpha_0(i))\) guarantees that \(F(\Theta_t) \to F^*\), completing the proof.
\end{proof}

\section{Proof of Theorem~\ref{thm:dynamic_regret}}
\label{app:dynamic_regret}

\begin{proof}
Recall the dynamic regret defined in~\eqref{eq:dynamic_regret}:
\[
R_T := \sum_{t=0}^{T-1} \left[ \mathcal{L}^{(t)}(\Theta_t) - \mathcal{L}^{(t)}(\Theta_t^*) \right],
\]
where \(\Theta_t^* := \arg\min_{\Theta \in \Omega} \mathcal{L}^{(t)}(\Theta)\) is the optimal trunk parameter at time \(t\), and let
\[
CV_T := \sum_{t=1}^{T-1} \|\Theta_t^* - \Theta_{t-1}^*\|.
\]

Under Assumption~\ref{assump:DR}, each \(\mathcal{L}^{(t)}\) is \(L\)-smooth and has gradients bounded by \(\|\nabla \mathcal{L}^{(t)}(\Theta)\| \le G\) for all \(\Theta \in \Omega\), and let \(D := \sup_{t,\Theta} \|\Theta_t - \Theta_t^*\|\). Then standard online convex analysis (cf.~\citep{bottou2018optimization,mairal2013stochastic}) yields:
\begin{equation*}
    \begin{aligned}
        \mathcal{L}^{(t)}(\Theta_t) - \mathcal{L}^{(t)}(\Theta_t^*) \le \frac{1}{2\alpha_0(t)} & \left( \|\Theta_t - \Theta_t^*\|^2 - \|\Theta_{t+1} - \Theta_t^*\|^2 \right)  \\
        & + \frac{\alpha_{L-1}(t)}{2} G^2 + \frac{D}{\alpha_0(t)} \|\Theta_t^* - \Theta_{t-1}^*\|.
    \end{aligned}
\end{equation*}
Summing over \(t = 0\) to \(T-1\) gives:
\[
\begin{aligned}
R_T &\le \frac{1}{2\alpha_0(0)} \|\Theta_0 - \Theta_0^*\|^2
+ \frac{G^2}{2} \sum_{t=0}^{T-1} \alpha_{L-1}(t)
+ D \sum_{t=1}^{T-1} \frac{\|\Theta_t^* - \Theta_{t-1}^*\|}{\alpha_0(t)} \\
&= \mathcal{O}\left( \sum_{t=0}^{T-1} \alpha_{L-1}(t) + \frac{CV_T}{\min_t \alpha_0(t)} \right),
\end{aligned}
\]
where \(CV_T = \sum_{t=1}^{T-1} \|\Theta_t^* - \Theta_{t-1}^*\|\) is the cumulative variation of the optimal path.

In particular, if \(\alpha_0(t) = \alpha_{L-1}(t) = \mathcal{O}(1/\sqrt{T})\), then:
\[
R_T = \mathcal{O}\left( \sqrt{T} + CV_T \cdot \sqrt{T} \right),
\]
which implies sublinear dynamic regret as long as \(CV_T = o(\sqrt{T})\).
\end{proof}

\newpage
\section{Visualization of Experiment on dataset}
We visualize the result from K-U-Net with 30 epochs adaptation. The left columns are training adaptation phase, and the right columns are testing phase. 
\label{appendix-figure-result}
\begin{figure}[htbp]
\centering
\includegraphics[width=0.9\linewidth]{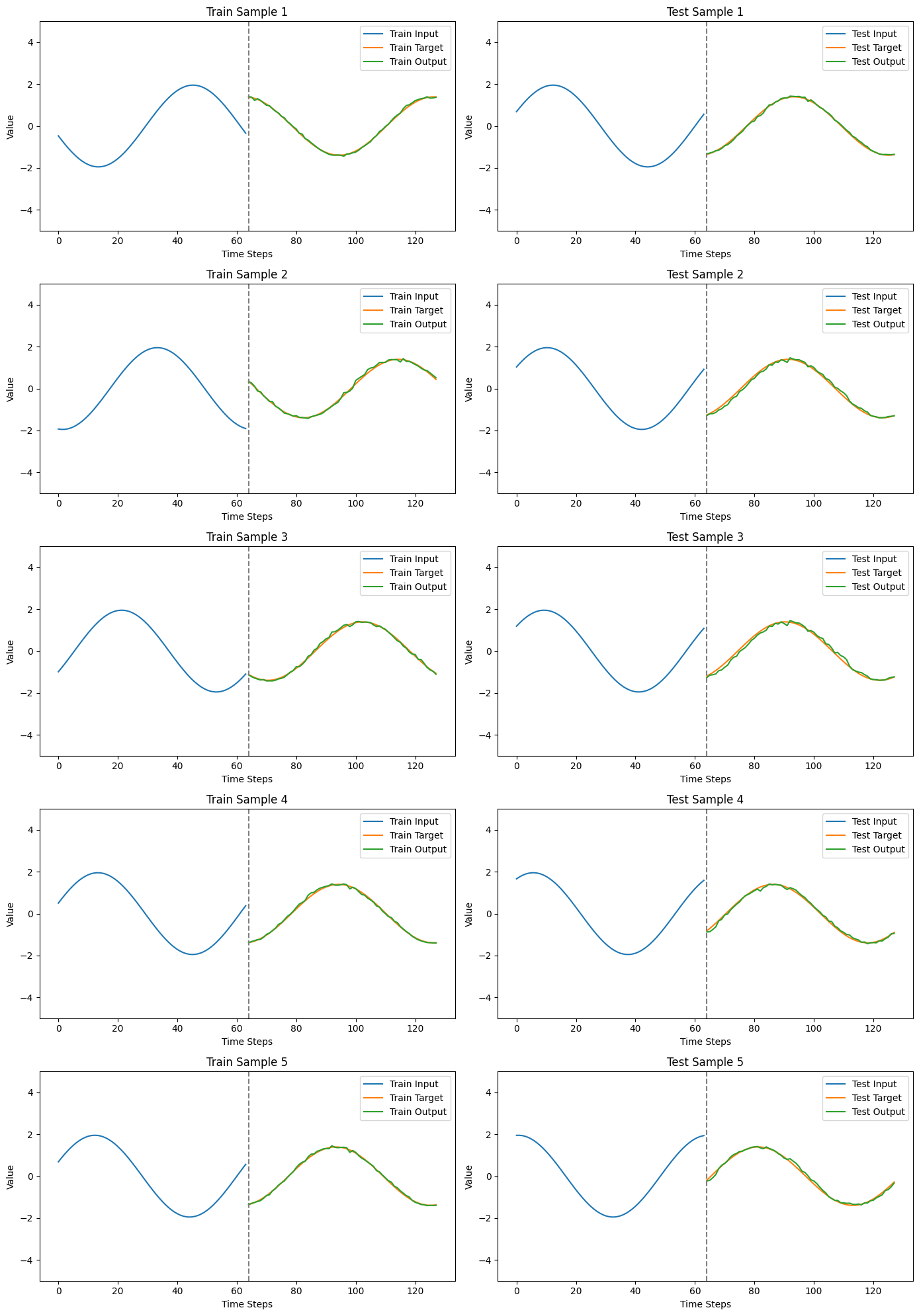}
\caption{
Prediction result of Dynamic K-U-Net on Dataset S1. The model consistently captures the target trajectories across all samples, validating its ability to adapt to structured task shifts with minimal error.
}
\label{fig:result-dataset-1}
\end{figure}

\begin{figure}[htbp]
\centering
\includegraphics[width=0.9\linewidth]{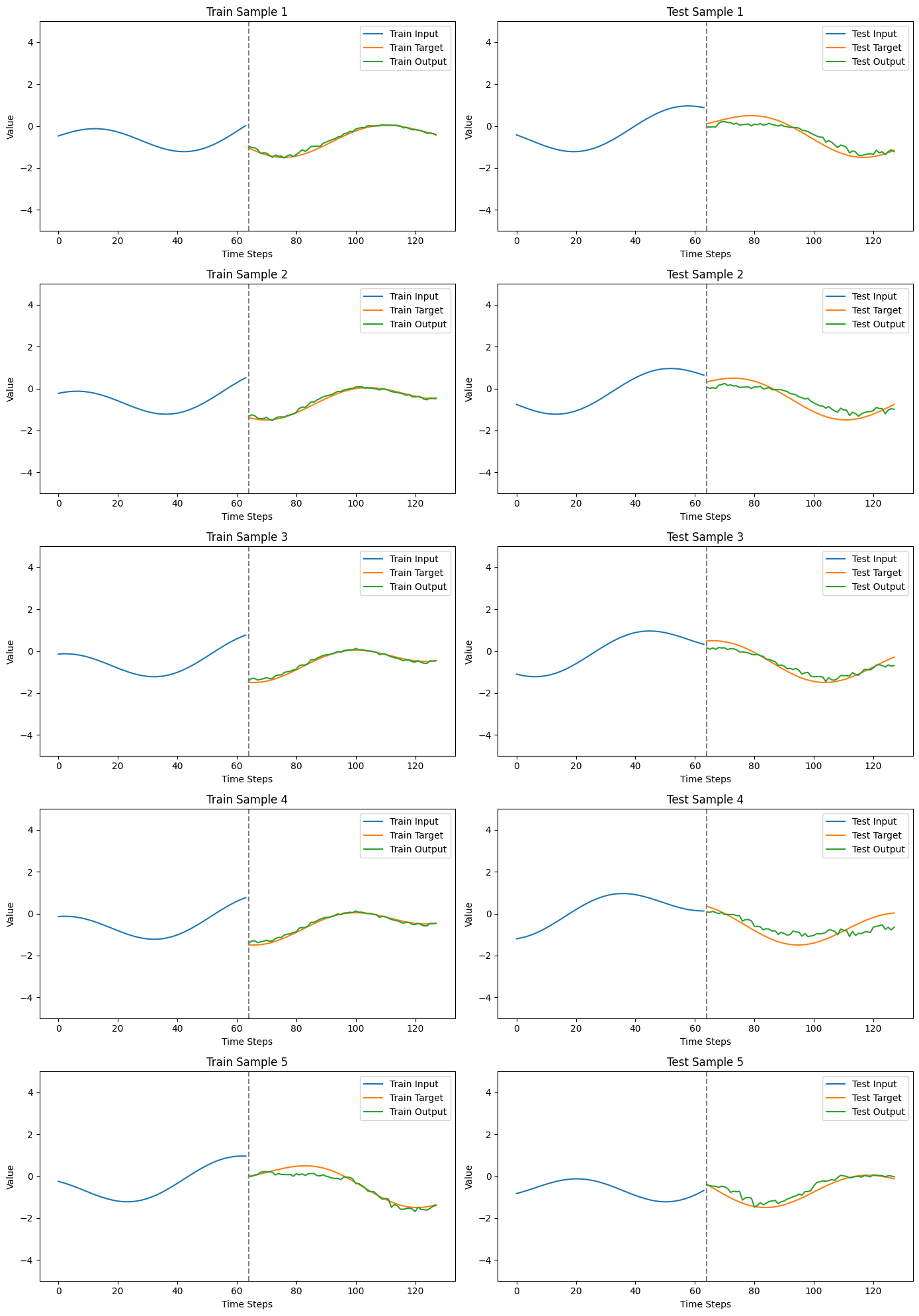}
\caption{
Prediction result of Dynamic K-U-Net on Dataset S2. Dynamic K-U-Net accurately tracks complex task mappings with asymmetric or partially conflicting objectives, demonstrating robust task-specific adaptation.
}
\label{fig:result-dataset-2}
\end{figure}

\begin{figure}[htbp]
\centering
\includegraphics[width=0.9\linewidth]{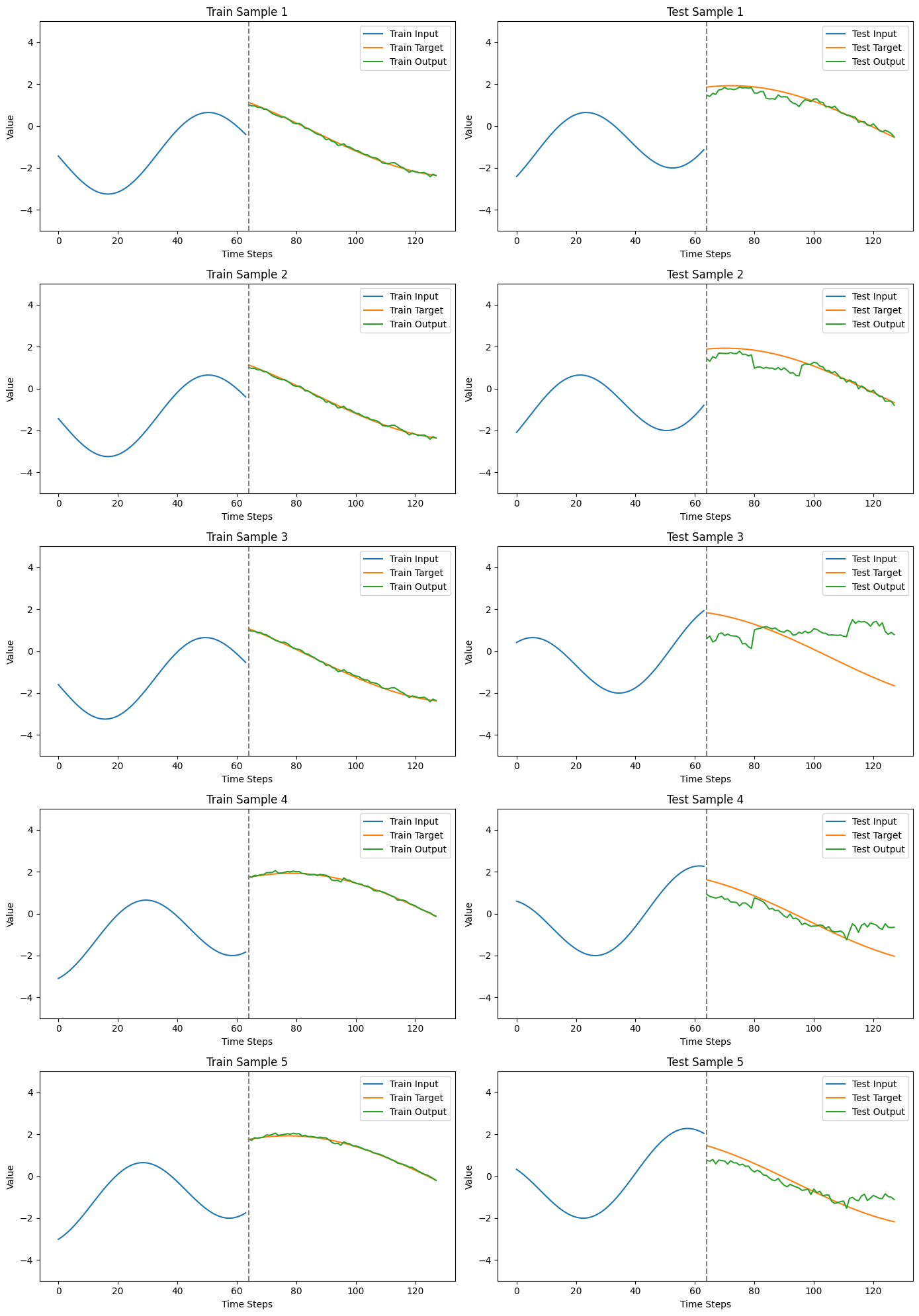}
\caption{
Prediction result of Dynamic K-U-Net on Dataset S3. Despite increasing ambiguity and overlap between subtasks, the model retains high predictive fidelity, highlighting its adaptability in dynamic and uncertain environments.
}
\label{fig:result-dataset-3}
\end{figure}

\end{document}